\documentclass[times, review, 10pt]{elsarticle}
\usepackage{amsmath,amsfonts}
\usepackage{algorithmic}
\usepackage{algorithm}
\usepackage{array}
\usepackage[caption=false,font=normalsize,labelfont=sf,textfont=sf]{subfig}
\usepackage{textcomp}
\usepackage{stfloats}
\usepackage{float}
\usepackage{url}
\usepackage{verbatim}
\usepackage{graphicx}
\journal{Pattern Recognition}
\begin{document}
\begin{frontmatter}

\title{PanoTPS-Net: Panoramic Room Layout Estimation via Thin Plate Spline Transformation}

\author{Hatem Ibrahem\textsuperscript{a}, Ahmed Salem\textsuperscript{b}, Qinmin Vivian Hu\textsuperscript{a}, Guanghui Wang\textsuperscript{a}}

\affiliation{organization={Department of Computer Science, Toronto Metropolitan University},
            addressline={350 Victoria St.}, 
            city={Toronto},
            postcode={M5B2K3}, 
            state={Ontario},
            country={Canada}}
\affiliation{organization={Department of Electrical Engineering, Faculty of Engineering, Assiut University},
            addressline={University street}, 
            city={Asyut},
            postcode={71515}, 
            state={Assiut},
            country={Egypt}}

\begin{abstract}
Accurately estimating the 3D layout of rooms is a crucial task in computer vision, with potential applications in robotics, augmented reality, and interior design. This paper proposes a novel model, PanoTPS-Net, to estimate room layout from a single panorama image. Leveraging a Convolutional Neural Network (CNN) and incorporating a Thin Plate Spline (TPS) spatial transformation, the architecture of PanoTPS-Net is divided into two stages: First, a convolutional neural network extracts the high-level features from the input images, allowing the network to learn the spatial parameters of the TPS transformation. Second, the TPS spatial transformation layer is generated to warp a reference layout to the required layout based on the predicted parameters. This unique combination empowers the model to properly predict room layouts while also generalizing effectively to both cuboid and non-cuboid layouts. Extensive experiments on publicly available datasets and comparisons with state-of-the-art methods demonstrate the effectiveness of the proposed method. The results underscore the model's accuracy in room layout estimation and emphasize the compatibility between the TPS transformation and panorama images. The robustness of the model in handling both cuboid and non-cuboid room layout estimation is evident with a 3DIoU value of 85.49, 86.16, 81.76, and 91.98 on PanoContext, Stanford-2D3D, Matterport3DLayout, and ZInD datasets, respectively. The source code is available at: \url{https://github.com/HatemHosam/PanoTPS_Net}.
\end{abstract}


\end{frontmatter}

\section{Introduction}
\label{intro}
Panoramic images are commonly used to solve various problems in computer vision, such as room layout estimation from a single image \cite{b0}. Estimating 3D room layouts from a single panorama image has emerged as a pivotal and promising subject with the advancement of computer vision and image processing technologies. The ability to automatically determine the spatial layout of walls, ceiling, floor, and other architectural features from a single, wide-angle snapshot has profound implications for interior design, virtual reality, robotics, computer animation, and augmented reality. This research leverages a combination of deep learning and geometric modeling, presenting the potential to revolutionize our interactions with the creation and interpretation of interior environments. 

Traditionally, defining the layout of a particular area requires costly physical measurements and architectural expertise. However, recent advances in computer vision and neural networks have automated this process. These technologies may deduce the three-dimensional structure and spatial layout of the environment by analyzing the visual clues presented in a single panoramic shot, thereby bridging the gap between the physical world and the virtual realms. Current deep learning-based methods for room layout estimation primarily rely on predicting the edge map mask or regressing the room corner points. This paper explores the intricacies of the room layout estimation task from a single panoramic image, investigating the possibility of utilizing spatial transformations, specifically the Thin Plate Spline (TPS) transformation, to address this challenge. Instead of dealing with the problem of room layout estimation as a semantic edge detection or a key point regression problem, we formulate it as an image warping problem by estimating the required TPS transformation parameters.

\begin{figure}[!t]
\centering
\includegraphics[width=0.85\linewidth]{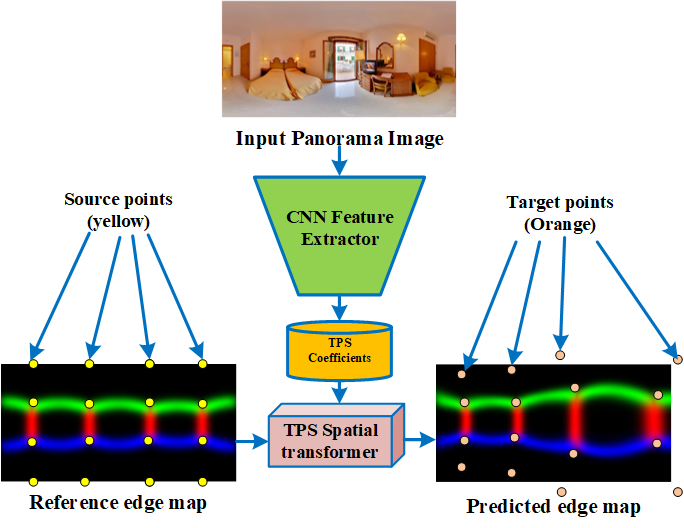}
\caption{Overview of the proposed method. The yellow points represent the source control point of the TPS transformation in the reference edge map, while the orange points represent the predicted control points in the target edge map. }
   \label{fig1}
\end{figure}

TPS transformation is a mathematical technique commonly used in image processing and computer graphics to smoothly and flexibly morph one shape into another. It is widely employed in image warping and mesh deformation. The expression ``thin plate" is obtained from bending a thin sheet of metal to accomplish the required change. The fundamental idea is to create a function that reduces bending energy while precisely transferring one set of control points to another, ensuring that nearby points in the source space remain close to each other in the target space. 

For each point pair ($x_i$, $y_i$) in the source point (the original point to be deformed) and ($x_i'$, $y_i'$) in the target point (the corresponding points after transformation), we aim to find a function $U(x, y)$ that maps ($x_i$, $y_i$) to ($x_i'$, $y_i'$). Here the TPS transformation $U(x, y)$ is defined as 
\begin{equation}
U(x, y) = a_0 + a_1 x + a_2 y + \sum(r_i^2 * log(r_i^2)) * b_i
\label{eq1}
\end{equation}
where $a_0$, $a_1$, $a_2$ are linear coefficients, $r_i$ is the Euclidean distance from $(x, y)$ to each source point $(x_i, y_i)$, $b_i$ is the weight for each source point. Note that $b_i$ determines the amount of influence of each source point on the transformation. These weights are calculated as
\begin{equation}
b_i =  \frac{\sum(r_i^2 * log(r_i^2))}{k} 
\label{eq2}
\end{equation}
where $k$ is a constant that governs the flexibility and smoothness of the transformation. A larger value of $k$ results in an abrupt and distorted transformation, while a smaller $k$ leads to higher flexibility. 
To determine the coefficients $a_0$, $a_1$, $a_2$, and $b_i$, one may solve a linear system of equations. In our approach, we employ a grid-based spatial transformation network (STN) \cite{b1} to learn those coefficients based on the features of the input panorama image. 

An overview of the proposed method is depicted in Fig. \ref{fig1}. The features of an input panorama image are extracted by a Convolutional Neural Network (CNN). These features are then utilized to predict the TPS transformation parameters. Subsequently, the TPS transformer layer is employed to deform the reference layout to the target layout. The TPS control points, highlighted in yellow in Fig. \ref{fig1}, are utilized by the TPS layer to deform the reference image according to \eqref{eq1} and \eqref{eq2} using a grid-based spatial transformer.

The main contributions of the paper are summarized below.
\begin{itemize}
\item We propose a novel method to estimate the room layouts using an image-warping technique, departing from the conventional methods of semantic edge detection and key point regression.
\item We design an end-to-end spatial transformation network architecture by incorporating the thin-plate spline transformation.
\item The proposed PanoTPS-Net can learn image warping in an unsupervised manner, eliminating the need for expensive warping annotations.
\end{itemize}

The rest of this paper is organized as follows. First, we discuss the related work in Section \ref{rel}. Then, we present the proposed method in detail in Section \ref{method}. After that, we present our experiments and results in Section \ref{exp}. We also present an ablation study in Section \ref{study}, and the conclusion including the limitations of our work and our future work in Section \ref{con}.

\section{Related work}
\label{rel}
Significant progress has been made in the field of 3D room layout estimation from a panoramic image in recent years. Hedau et al. \cite{a5} presented one of the earliest approaches for panoramic room layout estimation by using a parametric 3D "box" to describe the global room space and iteratively localize clutter while refitting the box to address the problem of crowded rooms. Zhang et al. \cite{b3} introduced PanoContext, one of the earliest frameworks for holistic 3D room layout estimation from a single panoramic image. The method assumes a Manhattan world structure and begins by estimating three orthogonal vanishing points directly from the equirectangular panorama. Using these vanishing directions, the model extracts structural lines and intersections to generate candidate cuboid layout hypotheses. Xu et al. \cite{a4} presented the Pano2CAD method for panorama image layout estimation by estimating positions and orientations of walls and objects using a Bayesian inference model. Zou et al. \cite{a8} proposed LayoutNet, a method for room layout estimation from panorama images by aligning the image based on the vanishing point. They designed a CNN to predict both the boundary map and corner map to well optimize the layout. 

Fernandez-Labrador et al. \cite{a10} introduced a technique that leverages panoramas to predict both the layout and bounding boxes of major items within a room. Their method is based on a simple 3D box representation of the room layout. They conducted experiments using the SUN360 and Stanford-2D3D datasets to demonstrate the advantages of estimating layouts by combining geometry and convolutional neural networks. A novel method for estimating the 3D room layout from a single panoramic image, known as HorizonNet, was presented by Sun et al. \cite{a3}. They encoded the room layout estimation problem as three 1D vectors representing wall-floor, wall-ceiling, and wall-wall corners. They also extended the Stanford 2D/3D dataset with annotated layouts for training and evaluation.  DuLa-Net, a dual-projection network developed by Yang et al. \cite{a7}, predicts room layouts from single RGB panoramas by utilizing equirectangular panorama-view and perspective ceiling-view. 
Wang et al. \cite{a1} defined the room layout estimation task as predicting the depth on a panorama's horizon line. They introduced LED2-Net, a monocular 360-degree layout estimate approach based on differentiable depth rendering. Extensive research on room layout estimation using panoramas was carried out by Xu et al. \cite{a6}. They employed a CNN to predict the layout of an input panorama image to guide another CNN for the view synthesis task. They also created an interactive application to facilitate the identification of 2D interior panoramas with high-quality 3D room plans. They later proposed an upgraded layoutNetV2 \cite{a8-1} as a common framework for layout estimation from panorama images based on pre-processing, layout prediction, and post-processing stages. 

Cruz et al. \cite{a8-3} proposed Zillow dataset for room layout estimation for empty spaces from panoramic images. The dataset contained empty spaces without any furniture to lower the complexity of the room layout estimation by introducing uncluttered scenes. 
Jiang et al. \cite{a2} proposed the LGT-Net, an indoor panoramic room layout estimation method that utilizes a geometry-aware transformer network. Shen et al. \cite{a0} proposed panoramic room layout estimation based on cross-scale distortion awareness. By pre-segmenting orthogonal (vertical and horizontal) planes from a complex scene, they were able to disentangle a 1D representation of the room and clearly identify the geometric clues needed for indoor layout estimation. These works show great advancements in 3D room layout estimation utilizing deep convolutional networks and panoramic cameras. Solarte et al. \cite{mlc1} proposed a self-training approach for panoramic room layout estimation using multi-view panoramas. 
Recently, Pu et al. \cite{b6} proposed Pano2Room, an automatic reconstruction of high-quality 3D indoor scenes from a single panoramic image. They constructed a preliminary mesh from the input panorama image and iteratively refined this mesh using RGBD images. Li et al. \cite{b78} proposed robust 3D point clouds classification based on declarative defenders. 



While the above-discussed methods have shown success in room layout estimation from panoramic images, our approach addresses the problem from a novel perspective by formulating it as an image-warping task. Unlike conventional methods that typically rely on semantic edge detection or keypoint regression, our method employs a simpler architecture yet achieves superior performance in most cases, as demonstrated through extensive experimental evaluations.

\begin{figure*}[!t]
\centering
  \centering
  \includegraphics[width=1\linewidth]{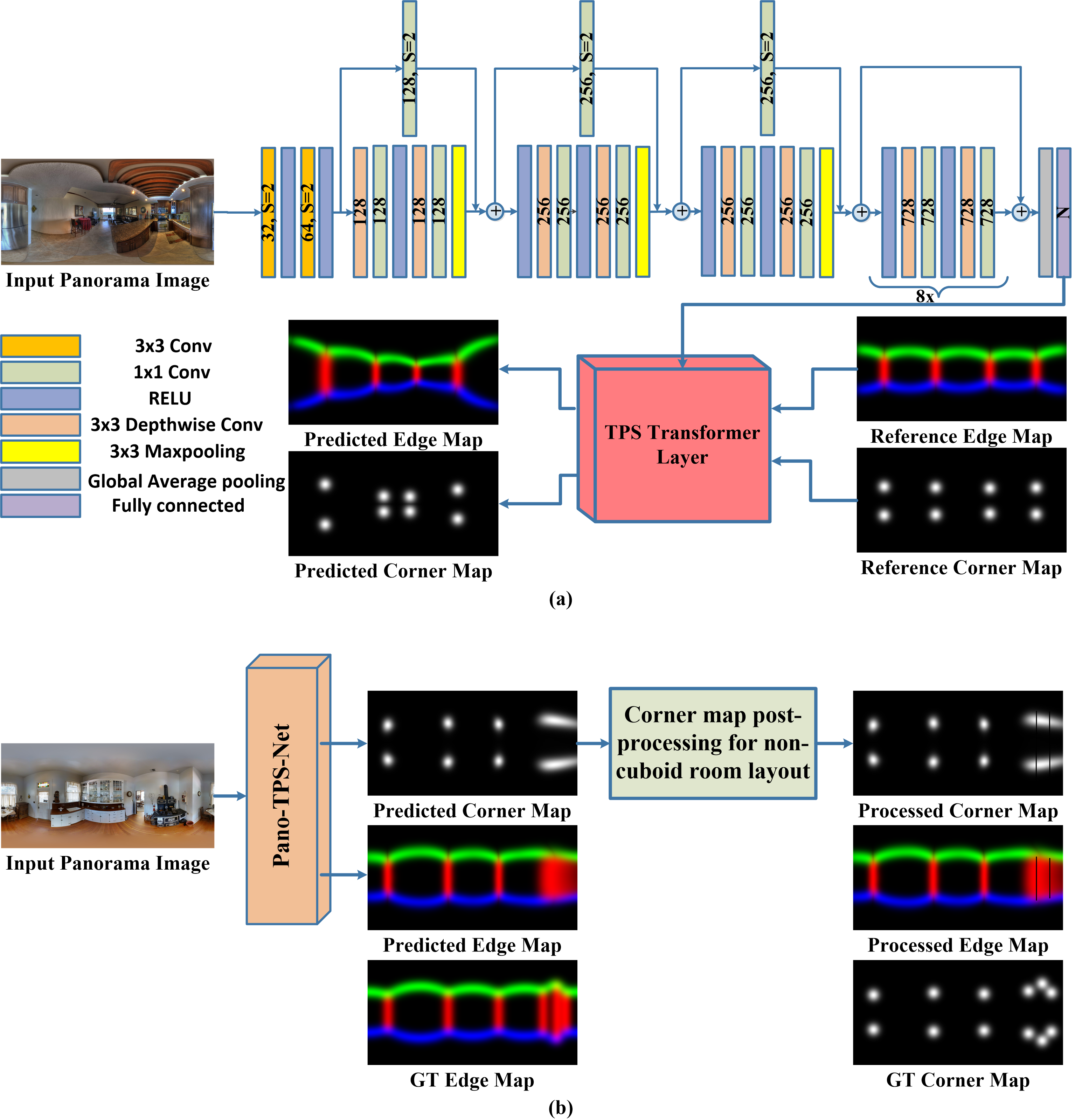}
   \caption{The proposed PanoTPS-Net architecture for 3D room layout estimation. (a) A modified Xception architecture (each layer shows the number of output features) is employed to extract the TPS transformation parameters, then those parameters are used by the TPS transformer layer to warp the reference layout (edge and corner maps) to the target layout. (b) The corner map post-processing is applied in the case of non-cuboid room layout estimation.}
   \label{fig2}
\end{figure*}

\section{Proposed method}
\label{method}
This section provides a detailed exposition of the two primary components of our proposed method for 3D room layout estimation: the feature extractor and TPS transformer layer. We also elucidate the training and test procedures.
The proposed PanoTPS-Net is illustrated in Fig. \ref{fig2}. It is composed of two major components: The feature extractor and the TPS transformer layer. This section provides a detailed description of the architecture.

\subsection{Feature extractor}
\label{feat}
The CNN feature extractor proposed depends mainly on the separable convolution in depth (DW-Conv2D) \cite{a11} in a form similar to the modified Xception architecture in \cite{a11-1} with some modifications. 
The proposed model consists of a sequence of standard convolutional layers (Conv2D) and DW-Conv2D with Relu activation, and max-pooling (only in the first three blocks). The DW-Conv2D consists of a depth-wise convolution (a $3\times3$ convolution through each channel separately) and a point-wise convolution (a $1\times1$ Conv2D). The DW-Conv2D is much more computationally efficient than Conv2D. The blocks' content is inherited from the original Xception architecture, but the number of blocks is fewer. Fig. \ref{fig2}-(a) shows the details of the architecture of the proposed feature extractor. After the feature extracting blocks, we add a Global Average pooling layer and a fully connected layer with $N$ neurons equivalent to the number of the control points for the TPS transformer. We have also conducted experiments with different numbers of control points in the TPS transformer in Section \ref{study}.

\begin{figure*}[!t]
\centering
  \centering
  \includegraphics[width=0.95\linewidth]{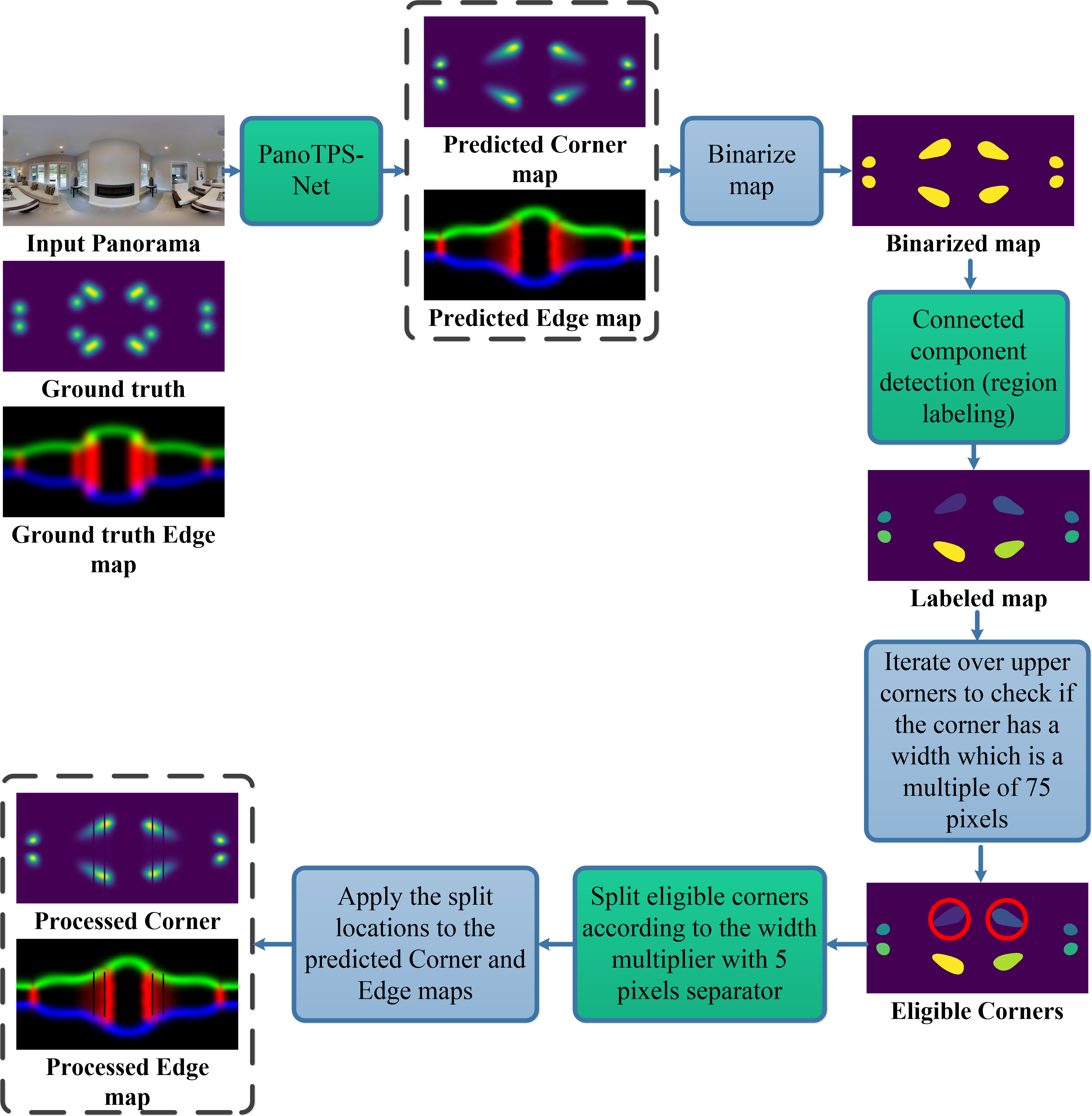}
   \caption{The proposed Corner map post-processing stage. This stage is applied in the case of non-cuboid room layout estimation. This stage includes passing the initially predicted corner map to a binarization stage, connected component detection, corner size analysis, and finally splitting corners into multiple corners based on a predefined criteria.}
   \label{fig2-a}
\end{figure*}

\subsection{Thin-plate spline transformer}
\label{TPS}
The spatial transformation network (STN) is a technique used in computer vision to perform differentiable linear and non-linear geometric transformations on images \cite{b1}. It is particularly useful for tasks such as image registration, image warping, and deformable image alignment. The underlying principle behind thin-plate splines is to find a flexible and smooth transformation that can map points from one form to another while minimizing the difference between the original and transformed images.

The thin-plate spline transformation can be represented in Eq. \eqref{eq3}.
\begin{equation}
T(\mathbf{X}) = \mathbf{A}\mathbf{X} + \mathbf{B} \mathbf{R}^2 \ln(\mathbf{R}^2)
\label{eq3}
\end{equation}
where $\mathbf{A}$ represents the linear coefficients $a_0$, $a_1$, and $a_2$ in Eq. \eqref{eq1}. They are normally set to one. $\mathbf{X}$ represents the interpolation grid points in the spatial transformer. $\mathbf{B}$ is the coefficients (i.e., control points) for the non-linear transformation. $\mathbf{R}$ is a matrix of pairwise distances between the points of the grid. Since we aim to minimize the error between the predicted and required edge and corner maps, we used a weighted sum of Huber's loss to learn the whole task. The Huber's loss is defined as.
\begin{equation}
L_\delta(y, y') = \begin{cases}
  \frac{1}{2}\sum_{p}(y - y')^2, & \text{for } |y - y'| \leq \delta, \\
  \delta(\sum_{p}|y - y'| - \frac{1}{2}\delta), & \text{otherwise}.
\end{cases}
\label{eq4}
\end{equation}
where $y$ denotes the ground truth pixel values. $y'$ represents the predicted pixel value. $p$ stands for the pixels in the edge map. $\delta$ is the threshold parameter (set to 1) that determines whether to use quadratic loss (MSE) or linear loss (MAE). Note that the non-linear parameters of TPS transformation $B$ are learned in an unsupervised way using the spatial transformer, which contains a grid used for the interpolation of the input reference edge map. The overall loss is calculated as a weighted sum of the two Huber losses.
\begin{equation}
L_{overall} =\alpha L_{edge-map} + \beta L_{corner-map}
\label{eq5}
\end{equation}
where the best results were obtained when $\alpha = 0.75$ and $\beta = 0.25$. We also trained the network with different values of $\alpha$ and $\beta$, and we report the evaluation results in Section \ref{study}.

\subsection{Corner map post-processing for non-cuboid layout estimation}
In the case of non-cuboid room layout estimation, we add an extra step to obtain a more accurate layout prediction. The initially predicted corner and edge maps combine close corners/edges in one larger corner/edge, resulting in incorrect shapes that exceed the normal edge or corner dimensions, as shown in the predicted edge/corner maps in Fig.\ref{fig2}-(b). We address this problem by employing a straightforward image processing technique. First, we binarize the corner map and employ connected component analysis to label all predicted corners with unique labels. Next, we iterate over the detected corners and check if the horizontal width of each corner falls within a predefined range (75 pixels in our experiments). We only process the upper corners of the walls since the lower corners are the corresponding corner points of the same edges. Then, we iterate over the upper corners and check the width of each one. If the width of any corner is a multiple of 75 pixels, we consider it in the splitting process. Additionally, we split the eligible corners into equal parts by a 5-pixel separator equivalent to the same multiples of the 75-pixel width which is shown as processed maps in Fig.\ref{fig2}-(b). Finally, we apply the same splitting to the predicted edge map. Fig. \ref{fig2-a} shows the details of the corner post-processing stage, including an example of a panorama image and the output of each step of the process. 

\subsection{Training and testing procedures}
During both the training and testing phases, we employ the feature extractor explained in Section \ref{feat} and the TPS transformer in Section \ref{TPS} back-to-back. In the training stage, we train the model to warp the reference layout (edge and corner maps) to the target layout. During testing, we utilize the trained model to warp the edge and corner map as shown in Fig. \ref{fig2}. We leverage both edge and corner maps simultaneously since they expedite the training process and yield better results compared to using the edge or corner map individually.
Our experiments show that warping both edge and corner maps results in better evaluation results than warping the edge or corner map separately. Comparisons between simultaneous and individual training using the edge and corner maps are reported in Section \ref{study}.

The training and testing processes were conducted on a desktop PC with Nvidia RTX-3090 GPU, Intel Core i7-8700 CPU @3.20 GHz, and 64 GB RAM. TensorFlow Keras was used to implement the model, and all models were initialized with ImageNet weights, and trained for 500 epochs. Adam's optimizer with weight decay was used for optimizing the model during training. All images were set to $1024\times512$ during training and testing.

\begin{figure*}
  \centering
  \includegraphics[width=1\linewidth]{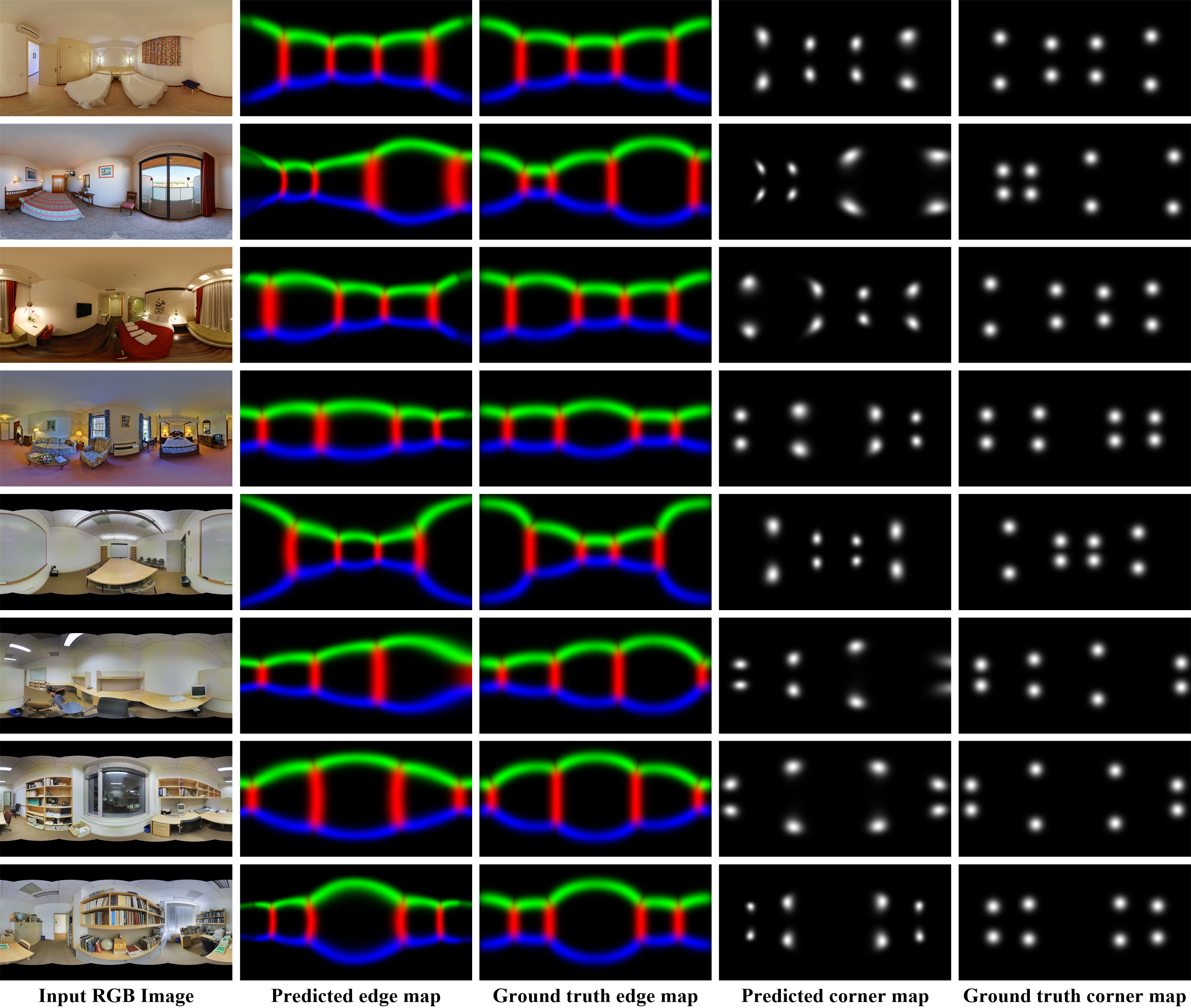}
   \caption{Sample results obtained by the proposed method. The top four rows represent sample results from PanoContext, and the bottom four rows represent sample results from Stanford-2D3D.}
   \label{fig3}
\end{figure*}

\section{Experimental results}
\label{exp}
In this section, we present the benchmarks used for training and testing, experiment design, results, and comparisons with the state-of-the-art (SOTA) methods in the room layout estimation task.

\subsection{Datasets for training and testing}
We train and test the proposed method on two widely used datasets for cuboid room layout estimation, Panocontext \cite{a8-1} and Stanford2D3D \cite{a8-1}, and two general room layout datasets (Matterport3DLayout and Zillow indoor dataset) that contain both cuboid and non-cuboid room layouts.

\textbf{PanoContext (PC)} is a popular dataset for room layout estimation from panorama images. It consists of 500 annotated indoor scenes of living rooms and bedrooms with cuboid layouts. We utilize the dataset split in \cite{a8} which splits the data into 90\% of the images for training and 10\% for testing.

\textbf{Stanford2D-3D (S2D3D)} is another dataset for room layout estimation from panorama images. The dataset originally consists of 1,413 panorama images of indoor scenes including offices, and classrooms without layout annotations. Zou et al. \cite{a8} provided a layout annotation for 571 images of the original dataset. PanoContext and Standford-2D3D have similar layout annotation done by Zou et al. \cite{a8} which consists of an RGB edge map (where red represents the wall-wall boundaries, green represents the wall-ceiling boundary, and blue represents the wall-floor boundary). The annotation also contains a grayscale map for corner point locations, and the numerical $x-y$ locations of the corners are also provided. We train and test our model on the annotated images from Stanford2D3D combined with the images of PanoContext with the same configurations and split provided in \cite{a8-1}.

\textbf{Matterport3DLayout (MP3D)} is a dataset derived from Matterport3D \cite{a8-2} by Zou et al. \cite{a8-1}. It contains both cuboid and non-cuboid room layouts with ground truth layout labels for edges and corners. It consists of 2,295 images with 1,647 training images, 190 validation images, and 458 test images, and the image size is $1024\times512$.

\textbf{Zillow indoor dataset (ZInD)} \cite{a8-3} is a large labeled dataset of panorama images of unfurnished residential spaces. It contains both cuboid and non-cuboid room layouts with the number of room corners ranging from 4 to over 10 corners per room. The dataset comprises 24,882 training images, 3,080 validation images, and 3,170 test images. For consistency, we resize all images to $1024\times512$.

\subsection{Evaluation metrics}
We employ four major evaluation metrics to measure the performance of the proposed method. The first metric is the 3D Intersection-over-Union (3DIoU) which measures the amount of intersection between the predicted 3D cuboid layout and the ground truth 3D cuboid layout over the union of the two cuboid layouts. Eq. \eqref{eq6} shows the formula of the 3DIoU metric.
\begin{equation}
3DIoU = \frac{Cube_{pred.} \cap Cube_{gt}}{Cube_{pred.} \cup Cube_{gt}}
\label{eq6}
\end{equation}

The second metric is the 2D Intersection-over-Union (2DIoU), which is similar to that in \eqref{eq6}. The only difference is that 2DIOU is used especially for the evaluation of the non-cuboid layouts by measuring the intersection-over-union of the predicted and ground truth layout areas instead of boxes. 
\begin{equation}
2DIoU = \frac{Area_{pred.} \cap Area_{gt}}{Area_{pred.} \cup Area_{gt}}
\label{eq6-1}
\end{equation}

The third metric is the Corner Error (CE) which measures the Euclidean distance between the location of the predicted corner point ($x_{pred}$, $y_{pred}$) and the location of the ground truth corner point ($x_{gt}$, $y_{gt}$). 
\begin{equation}
CE = \sum_{i=1}^{C}\sqrt{(x_{pred} - x_{gt})^2 + (y_{pred} - y_{gt})^2}
\label{eq7}
\end{equation}
where $i$ is an iterator over $C$ corner points. The fourth metric is the pixel error (PE). PE measures the difference between the predicted pixels ($p_i$) and ground truth pixels ($g_i$) of the edge map (which contains the ceiling-wall, wall-wall, and wall-floor boundaries) as in Eq. \eqref{eq8}.
\begin{equation}
PE = \frac{1}{N}\sum_{i=1}^{N}\delta(p_i,g_i)
\label{eq8}
\end{equation}
where $i$ is an iterator over $N$ pixels in the edge map. $\delta$ is a function that measures the difference between $p_i$ and $g_i$.
\begin{table}[hbt!]
  \centering
   \caption{Quantitative comparison on PanoContext and Stanford-2D3D datasets for the cuboid room layout estimation task with SOTA methods. Setting 1 refers to training on PanoContext + whole Stanford2D-3D while testing on test samples from PanoContext, and Setting 2 refers to training on Stanford2D-3D + whole PanoContext while testing on test samples from Stanford 2D-3D.}
\label{tab1}
  \begin{tabular}{@{}llclc}
    \hline
    Dataset&Method & 3DIoU(\%)& CE(\%)&  PE(\%) \\
    \hline
    &Dula-Netv2 \cite{a8-1} & 83.77& 0.81 & 2.43 \\
    &LayoutNetv2 \cite{a8-1} & 85.02 & \bf{0.63} & \bf{1.79} \\
    &DMH-Net \cite{b4} & 85.48 & 0.73 & 1.96 \\
    &HorizonNet \cite{a3} & 82.63 & 0.74 & 2.17 \\
    Setting 1\;\;&LED2Net \cite{a1} & 82.75 & - & - \\
    &LGT-Net \cite{a2} & 85.16 & -  &- \\
    &LGT-Net+ Post-proc \cite{a2} & 84.94 & 0.69 & 2.07\\
    &DOPNet \cite{a0} & 85.46 &-& -\\
    &DOPNet+Post-proc \cite{a0} & 85.00 & 0.69 & 2.13 \\
    &\bf{Ours} & \bf{85.49}& -& - \\
    &\bf{Ours+Post-proc} & 84.94& \bf{0.63}& 2.06 \\
    \hline
    &Dula-Netv2 \cite{a8-1} & \bf{86.60}& 0.67 & 2.48 \\
    &LayoutNetv2 \cite{a8-1} & 82.66 & 0.83 & 2.59 \\
    &DMH-Net \cite{b4} & 84.93 & 0.67 & 1.93 \\
    &HorizonNet \cite{a3} & 82.72 & 0.69 & 2.27 \\
    &LED2Net \cite{a1} & 83.77 & - & - \\
    Setting 2\;\;&AtlantaNet \cite{a13} & 83.94 & 0.71 & 2.18 \\
    &LGT-Net \cite{a2} & 85.76 & -  &- \\
    &LGT-Net+Post-proc \cite{a2} & 86.03 & 0.63 & 2.11\\
    &DOPNet \cite{a0} & 85.47 &-& -\\
    &DOPNet+Post-proc \cite{a0} & 85.58 & 0.66 & 2.10 \\
    &Shape-Net \cite{a30} & 86.19 & 0.63 &1.90 \\
    &\bf{Ours} & 86.18& -& - \\
    &\bf{Ours+ Post-proc} & 85.92& \bf{0.62}& \bf{1.91} \\
    \hline
  \end{tabular}
\end{table}
\subsection{Evaluation results on cuboid room layouts}
\label{results}
We train the proposed model on cuboid room layout estimation with two configurations following \cite{a8-1}. In the first configuration (setting 1), we train the model on PanoContext training set + whole Stanford-2D3D, and then we evaluate on PanoContext test set. In the second configuration (setting 2), we train on Stanford-2D3D train set + whole PanoContext, then we evaluate on Stanford-2D3D test set. We followed the same preprocessing and post-processing scheme proposed by Zou et al. \cite{a8-1}. The obtained 3DIoU, CE, and PE using the two configurations on PanoContext and Stanford-2D3D are shown in Table \ref{tab1}. The proposed model with the first training configuration could achieve a 3DIoU of 85.49 and with post-processing \cite{a8-1}, it achieved a 3DIoU of 84.94, CE of 0.63 and PE of 2.06 outperforming the SOTA methods in terms of both 3DIoU and PE. With the second training configuration, our method could achieve a 3DIoU of 86.18, and with postprocessing \cite{a8-1}, it could achieve a 3DIOU of 85.92, CE of 0.62, and PE of 1.91, outperforming the SOTA methods in terms of CE and PE. Those results outperform most of the SOTA methods with a much simpler architecture than others. Fig.\ref{fig3} shows some sample qualitative results obtained by the proposed method on PanoConext and Stanford-2D3D datasets. 

\begin{figure*}[t]
  \centering
  \includegraphics[width=1\linewidth]{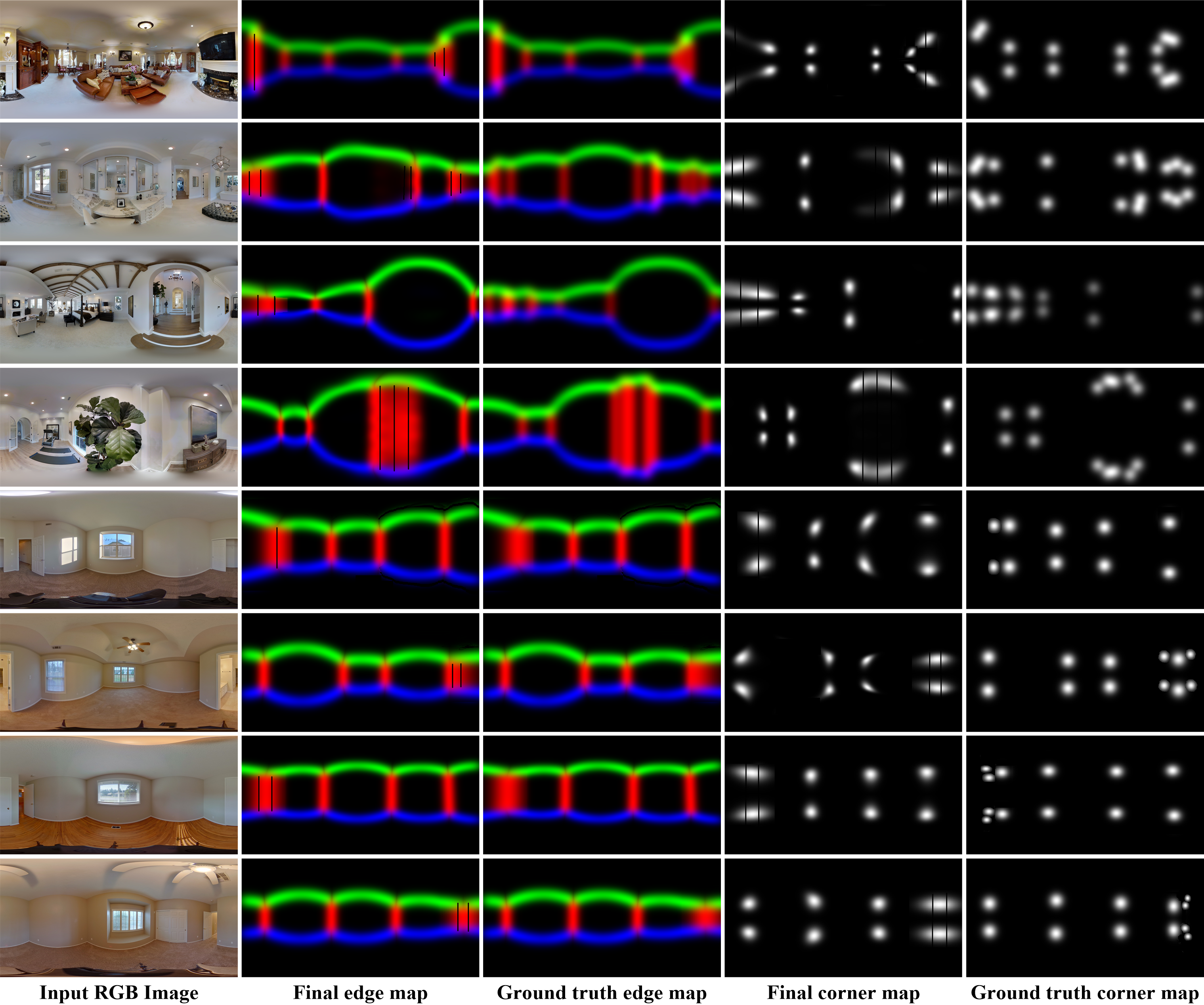}
   \caption{Sample results obtained by the proposed method on Matterport3DLayout (the top 4 rows) and Zillow indoor dataset (the bottom four rows) for non-cuboid room layout estimation. The results show the final corner/edge maps after corner map post-processing shown in Fig.\ref{fig2-a}}
   \label{fig3-1}
\end{figure*}

\begin{figure*}[t]
  \centering
  \includegraphics[width=0.9\linewidth]{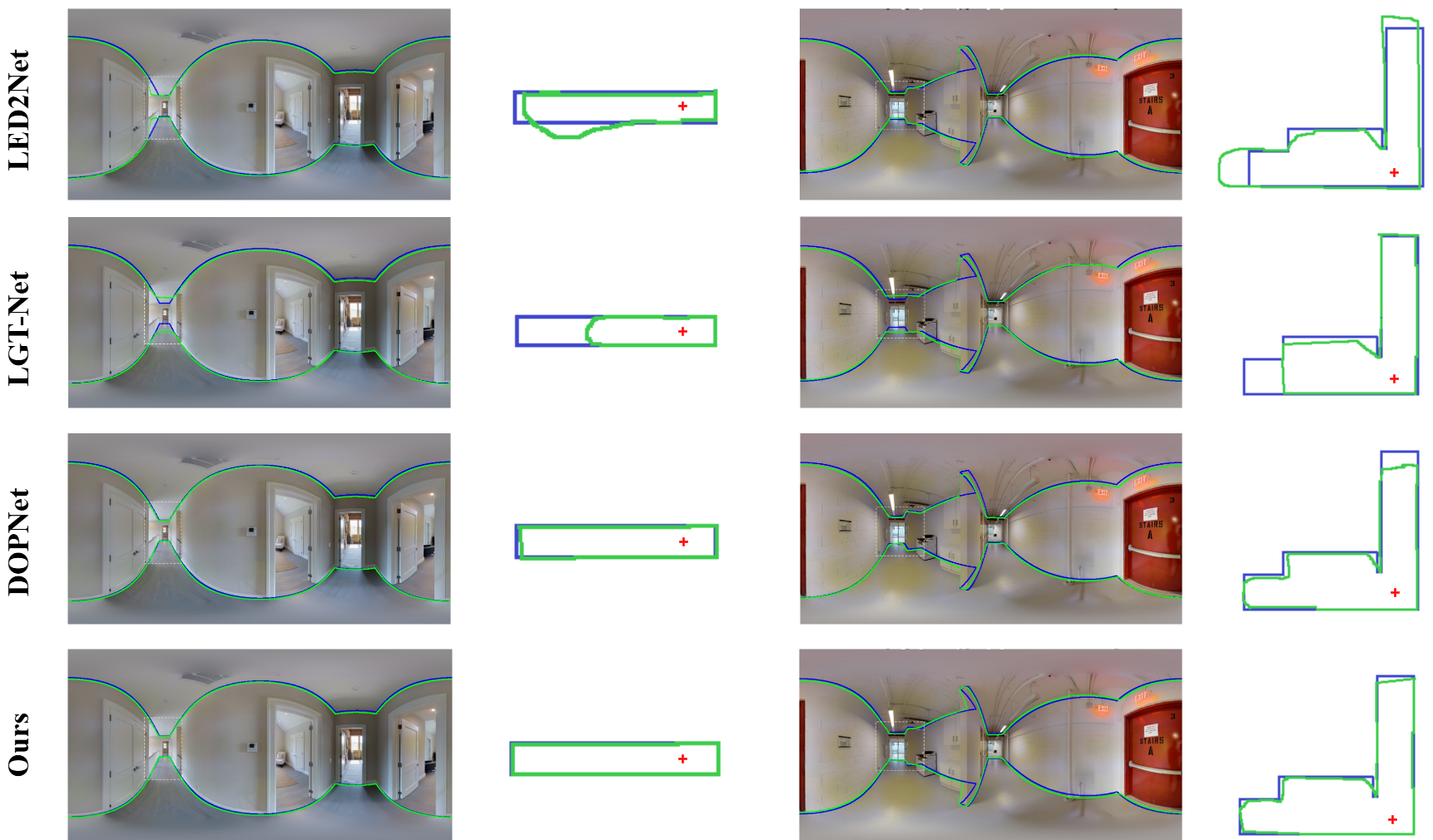}
   \caption{Qualitative comparison of the room layout estimation between LED2Net \cite{a1}, LGT-Net \cite{a2}, DOPNet \cite{a0}, and the proposed PanoTPS-Net. The ground truth layout, predicted layout, and camera position are shown in blue, green, and red, respectively.}
   \label{fig5}
\end{figure*}

\begin{figure*}[t]
  \centering
  \includegraphics[width=0.9\linewidth]{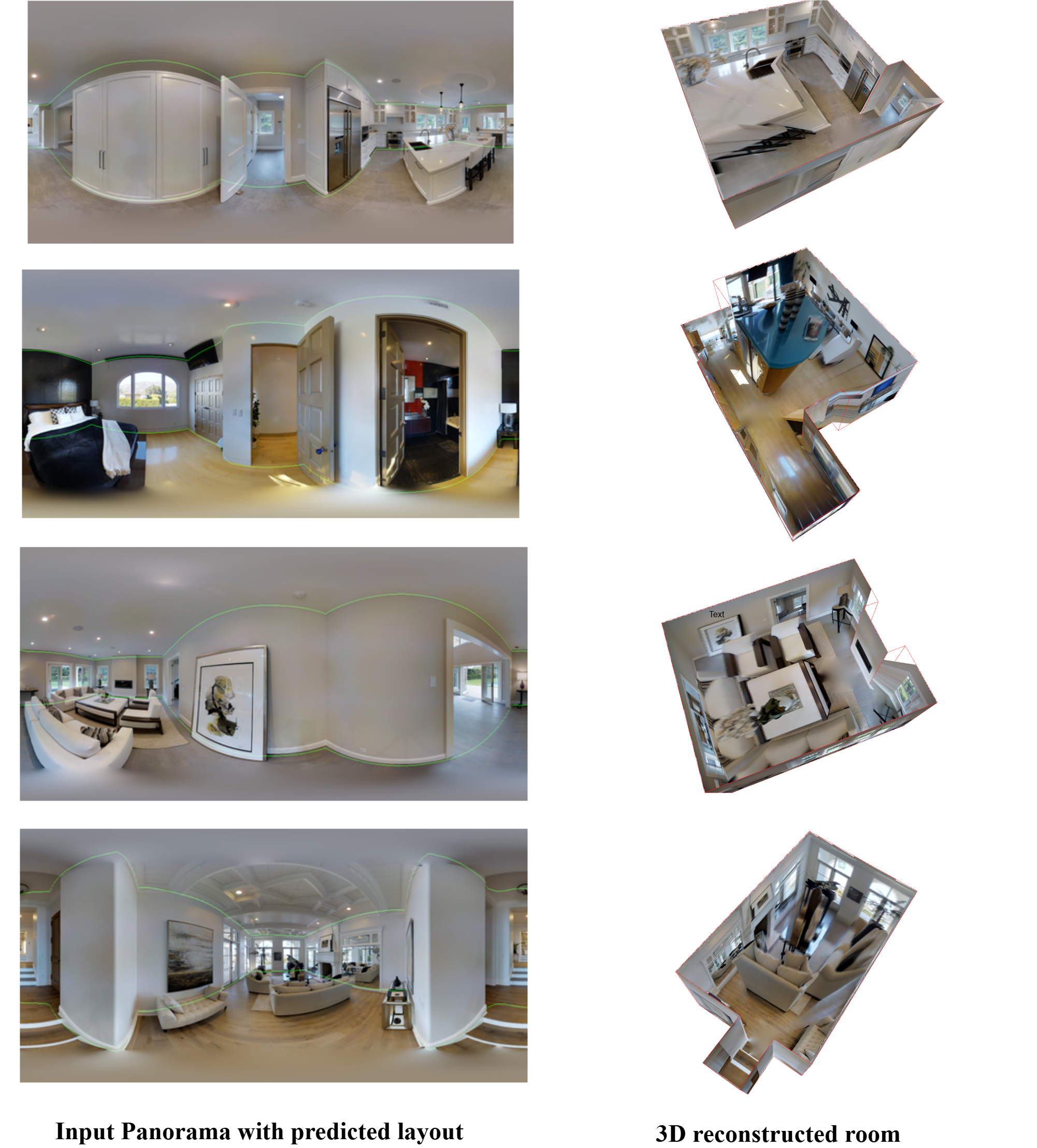}
   \caption{3D reconstruction results based on the proposed method on the Matterport3D-Layout dataset.}
   \label{fig6}
\end{figure*}

\subsection{Evaluation results on non-cuboid room layouts}
We trained the proposed method on Matterport3DLayout and ZInD datasets for the task of general room layout estimation (including both cuboid and non-cuboid layouts). To enhance accuracy, we applied corner map post-processing to refine the initial edge/corner map predictions. Fig. \ref{fig3-1} presents four sample results produced by PanoTPS-Net on Matterport3D. Here, the final edge/corner maps denote those obtained after applying corner map post-processing.
Without relying on panorama post-processing, our method achieves superior performance compared to the SOTA, with a 3DIoU of 81.76\% and 2DIoU of 84.15\% on Matterport3DLayout and a 3DIoU of 91.98\% and a 2DIoU of 90.05\% on ZInD, as shown in Table \ref{tab1-1}.

We further present a qualitative comparison between the proposed method and three SOTA methods (LED2Net \cite{a1}, LGTNet \cite{a2}, and DOPNet \cite{a0}) of room layout estimation in bird’s-eye view (i.e., top view) in Fig. \ref{fig5}. Specifically, we follow the equirectangular projection model, where the horizontal field of view (FOV) is 360\textdegree and the vertical FOV is 180\textdegree. Additionally, Fig. \ref{fig6} presents 3D reconstruction examples of rooms based on the predicted layouts, assuming a fixed camera height of 1.6 $m$ for scene reconstruction. 

The proposed PanoTPS-Net model contains 22.3 million parameters and achieves a runtime of approximately 200 milliseconds (5 frames per second). The corner post-processing step requires an additional 150 milliseconds, resulting in an overall inference time of about 350 milliseconds on an RTX 3090 GPU.

\begin{table}[hbt!]
  \centering
  \caption{Quantitative comparison on Matterport3DLayout and ZInD datasets for the non-cuboid room layout estimation task with SOTA methods.}
  \label{tab1-1}
  \begin{tabular}{@{}lclc}
    \hline
    Dataset &Method & 3DIoU(\%)& 2DIoU(\%) \\
    \hline
    &Dula-Netv2 \cite{a8-1} & 75.05& 78.73 \\
    &LayoutNetv2 \cite{a8-1} & 75.82& 78.73 \\
    &DMH-Net \cite{b4} & 78.97 & 81.25  \\
    &HorizonNet \cite{a3} & 79.11 & 81.71 \\
    &AtlantaNet \cite{a13} & 80.02 & 82.09 \\
    Matterport3D&HoHoNet \cite{a19-1} & 79.88 & 82.32 \\
    &PSMNet \cite{a19-2} & - & 81.01 \\
    &LED2-Net \cite{a1} & 81.52 & 83.91 \\
    &LGT-Net \cite{a2} & 81.11 & 83.52 \\
    &Shape-Net \cite{a30} & 81.2 & 83.93 \\
    &DOPNet \cite{a0} & 81.70 & 84.11\\
    &\bf{Ours} & \bf{81.76}& \bf{84.15} \\
    \hline
    &HorizonNet\cite{a3} & 90.44 & 88.59 \\
    ZInD&LED2-Net \cite{a1} & 90.36 & 88.49 \\
    &LGT-Net \cite{a2} & 91.77 & 89.95 \\
    &DOPNet \cite{a0} & 91.94 & \bf{90.13}\\
    &\bf{Ours} & \bf{91.98}& 90.05 \\
    \hline
  \end{tabular}
\end{table}

\section{Ablation study}
\label{study}
We conducted various studies on the feature extractor, the map used for warping, the weight values in the loss function, and the number of control points.

\begin{table}[hbt!]
  \centering
 \caption{Comparison between different feature extractors in training on PanoContext + Whole Stanford-2D3D.}
  \label{tab7}
  \begin{tabular}{@{}lclc}
    \hline
    Feature extractor & convergence & 3DIoU(\%) \\
    \hline
    ResNet50 \cite{a15} & weak& 31.44 \\
    ResNet50v2 \cite{a16} & weak& 37.12 \\
    Inceptionv3 \cite{a17} & weak & 42.73 \\
    DenseNet-121 \cite{a19} & weak & 42.92 \\
    MobileNetv3 \cite{a27} & Yes & 45.57 \\
    EfficientNetB1 \cite{a21} & Yes& 53.33\\
    EfficientNetB2 \cite{a21} & Yes & 57.89\\
    ConvNext-tiny \cite{a18} & Yes & 65.52\\
    ConvNext-small \cite{a18} & Yes & 74.24\\
    ViT-B/16 \cite{a29} &Yes& 81.61 \\
    \bf{MXception} & Yes & \bf{85.49} \\
    \hline
  \end{tabular}
\end{table}

\textbf{Feature extractor architecture}. We trained various feature extractor networks for our model. We trained common CNN architectures on the task of panorama layout estimation using our approach. We trained ResNet50 \cite{a15}, ResNet50v2 \cite{a16} InceptionV3 \cite{a17}, ConvNext-Tiny \cite{a18}, ConvNext-small \cite{a18} DenseNet-121 \cite{a19}, Efficient-NetB1 \cite{a21}, Efficient-NetB2 \cite{a21}, ViT-B \cite{a29} and our main architecture (MXception). Most of the architectures were not aligned well with the task and the loss did not improve during training. It was fixed in a large loss value. Table \ref{tab7} shows the training trials using previous feature extractors. Feature extractors such as ResNet50, Inceptionv3, DenseNet121, ConvNext-tiny, and EfficientNetB1 struggled to fit the task, with slow improvement in the loss. An architecture such as ViT-B attains good 3DIoU values but still MXception attains the best 3D-IOU and has much lower complexity than ViTs. 
Based on the experimental results, we choose MXception over other feature extractors.

\begin{table}[hbt!]
  \centering
\caption{Comparison between different Xception-like architectures with different network depths. The feature extractors with different layer depths were trained on PanoContext + Whole Stanford-2D3D. Note that layer names are the names used in the original Xception architecture, and global average pooling and dense layers are added after the layer shown in each row.}
  \label{tab8}
  \begin{tabular}{@{}lclc}
    \hline
    Output layer in Xception & 3DIoU(\%) \\
    \hline
    block7\_sepconv2\_act & 62.74 \\
     block10\_sepconv2\_act & 67.42 \\
     block13\_sepconv2\_act (MXception) & \bf{85.49}\\
    block14\_sepconv2\_act (Xception) & 79.64 \\
    \hline
  \end{tabular}
\end{table}

\textbf{Depth of the feature extractor}. We conducted a study to determine the optimal depth of the feature extractor. Starting with the original Xception architecture, we tried to train our model with a feature extractor taking the output from block\_7\_sepconv2\_act, block\_10\_sepconv2\_act, block\_13\_sepconv2\_act, block\_14\_sepconv2\_act (Original Xception architecture) layers. In each case, a global average pooling and a dense layer with neurons equal to double the control points are added after the output layer defined in each row of the table. Table \ref{tab8} shows a comparison between the 3DIoU in each case where the output taken from block\_13\_sepconv2\_act layer (which is MXception architecture) achieves the best performance.

\begin{table}[hbt!]
  \centering
\caption{Comparison between different configurations of the map used for TPS warping. The converge column shows if using the listed map makes the model converge during the training process. The 3DIoU values were measured on PanoContext (PC) and Matterport3D (MP3D).}
  \label{tab5}
  \begin{tabular}{@{}lclc}
    \hline
    \bf{Dataset}& &\quad\bf{PC} & \bf{MP3D} \\
    \hline
    Map to warp & Convergence & 3DIoU(\%) & 3DIoU(\%) \\
    \hline
    Corner map & Weak &  46.64 & 41.47\\
    Edge map & Yes& 82.71 & 78.11\\
    Corner and edge maps & Yes& \bf{85.49} & \bf{81.76}\\
    \hline
  \end{tabular}
\end{table}

\begin{table}[hbt!]
  \centering
 \caption{Comparison between using different weight values of $\alpha$ (edge map loss weight) and $\beta$ (corner map loss weight). In each case, the model was evaluated by the 3DIoU on PanoContext (PC) and Matterport3D (MP3D).}
  \label{tab6}
  \begin{tabular}{@{}lclc}
    \hline
    \bf{Dataset}&& \quad\bf{PC} & \bf{MP3D} \\
    \hline
    Edge map ($\alpha$) & Corner map ($\beta$) & 3DIoU(\%) & 3DIoU(\%)\\
    \hline
    \quad 0.10 & 0.90 & \quad 32.14 & 28.45\\
    \quad 0.25 & 0.75& \quad 67.71 & 60.12\\
    \quad 0.50 & 0.50& \quad 82.63 & 75.74\\
    \quad 0.75 & 0.25& \quad \bf{85.49}& \bf{81.76} \\
    \quad 0.90 & 0.10& \quad 83.95 & 77.28 \\
    \hline
  \end{tabular}
\end{table}

\textbf{Warping the edge map vs. corner map vs. both edge and corner maps}. We performed experiments to identify the optimal configuration for training the proposed model either using the edge map only, the corner map only, or both maps. Training the model solely with the corner map did not lead to convergence during training. However, when using the edge map alone and both the corner and edge in separate training experiments, the model converges successfully. Table \ref{tab5} shows that training with both maps demonstrated better evaluation value (3DIoU of 85.49\% on PC and 81.76\% on MP3D) than using the edge map only (3DIoU of 82.71\% on PC and 78.11\% on MP3D).
\begin{table}[hbt!]
  \centering
 \caption{Comparison between using different numbers of control points in the TPS transformer layer. In each case, the model was evaluated in terms of 3DIoU on PanoContext (PC) and Matterport3D (MP3D), respectively.}
  \label{tab7}
  \begin{tabular}{@{}lclc}
    \hline
     \bf{Dataset}& & \quad\bf{PC} & \bf{MP3D}\\
    \hline
    No. of control points & Convergence & 3DIoU(\%)& 3DIoU(\%)\\
    \hline
    \quad\quad\quad ($2\times2$) 4 & No & \quad - &\quad - \\
    \quad\quad\quad ($3\times3$) 9 & Yes & \quad35.94 & \quad27.22 \\
    \quad\quad\quad ($4\times4$) 16 & Yes & \quad \bf{85.49} & \quad 78.34\\
    \quad\quad\quad ($5\times5$) 25 & Yes & \quad 77.57 & \quad 79.71 \\
    \quad\quad\quad ($6\times6$) 36 & Yes & \quad 84.14 & \quad 79.94 \\
    \quad\quad\quad ($7\times7$) 49 & Yes & \quad 81.47 & \quad 80.09 \\
    \quad\quad\quad ($8\times8$) 64 & Yes & \quad 84.01 & \quad  \bf{81.76}\\
    \quad\quad\quad ($9\times9$) 81 & Yes & \quad 85.04 & \quad 80.84 \\
    \hline
  \end{tabular}
\end{table}

\textbf{Weights ratio in the loss function}. We trained the proposed model with different weight values ($\alpha$ and $\beta$) in the loss function defined in Eq. \eqref{eq5}. We tried the loss function with different mixing values (0.10, 0.25, 0.50, 0.75, and 0.90) and reported the 3DIoU in Table \ref{tab6}. The results show that the edge map has a significantly positive effect on the convergence of the loss. Using a higher value of $\beta$ and a lower value of $\alpha$ yields lower 3DIoU, as shown in the first and second rows of Table \ref{tab6}. However, employing higher $\alpha$ values than $\beta$ values results in improved convergence of the loss, specifically, setting $\alpha=0.75$ and $\beta=0.25$ yields optimal weight values, enabling the model to achieve the best 3DIoU values on both PC and MP3D. The reason is that the edge map is a dense map with three channels (RGB map), while the corner map is a sparse single-channel map with limited information about the layout.

\textbf{Optimal number of the TPS control points}. We conducted a study on the crucial parameter in the TPS transformer layer, the number of control points. We trained our model utilizing the MXception feature extractor with 4, 9, 16, 25, 36, 49, 64, and 81 control points in the TPS layer to decide the optimal number of control points. Note that the TPS transformer utilizes a square-shaped grid of control points. Our findings revealed that a small number of control points, such as 4 or 9 points, were insufficient to effectively learn the transformation task due to limited transformation flexibility. The optimal number of points that enable the model to achieve the best performance (3DIoU=85.49) is 16 points for the PC dataset and 64 points for the MP3D, as depicted in Table \ref{tab7}. Using larger numbers of control points shows a degradation in the performance at the cuboid layout estimation because the model excessively uses the control points to deform the reference layout leading to the prediction of more corners/edges than those for the target layout, and it shows an improvement in the performance in the non-cuboid layout estimation due to the need for more corners/edges than those exist in the reference layout to achieve to the target layout.

\begin{figure*}
\centering
  \centering
  \includegraphics[width=1\linewidth]{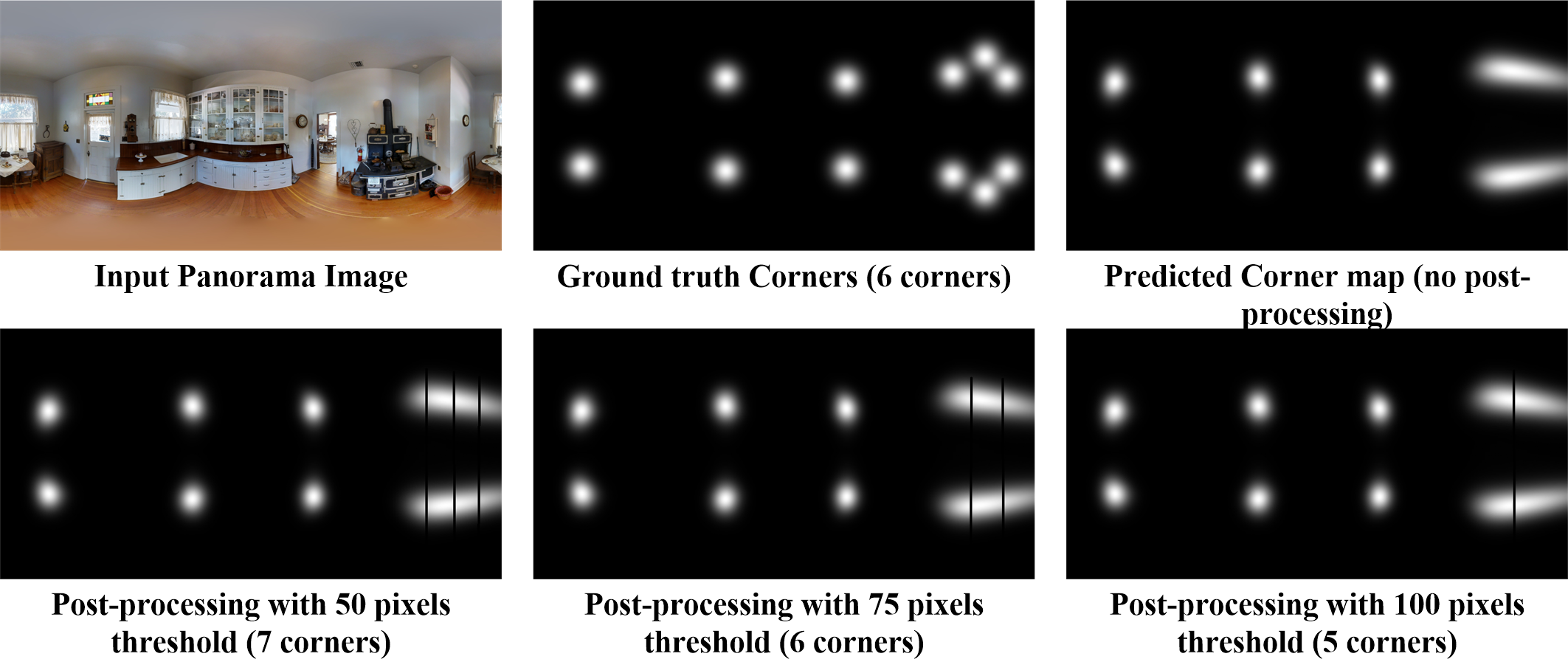}
   \caption{Visual comparison of the 50-, 75-, and 100-pixel thresholds used in corner-splitting post-processing for non-cuboid room layout estimation. The 75-pixel threshold yields the best performance, as it produces the same number of corners as in the ground truth corner map. }
   \label{fig7}
\end{figure*}

\textbf{Splitting threshold in non-cuboid layout post-processing}. 
We conducted a study to determine the optimal pixel threshold for splitting merged corner regions initially predicted by PanoTPS-Net. As shown in Fig. \ref{fig7}, we evaluated thresholds of 50, 75, and 100 pixels. The 75-pixel threshold produced the best results, yielding the same number of corners as in the ground truth corner map (6 corners). This aligns with the average corner diameter in the Matterport-Layout dataset at a resolution of $1024 \times 512$, which is approximately 75 pixels with a standard deviation of 5 pixels.

\section{Conclusion}
\label{con}
This paper has presented a novel and effective approach, PanoTPS-Net, for the task of 3D room layout estimation from single panorama images. The method leverages a CNN in conjunction with a TPS transformation, demonstrating a distinctive and robust synergy between these components. 
The architecture is structured in two carefully designed stages. In the first stage, a CNN extracts high-level spatial features from the input panorama. In the second stage, a TPS transformation layer warps a reference layout based on predicted parameters, enabling accurate and generalizable layout predictions.

Comprehensive evaluations were conducted on three widely used datasets, with PanoTPS-Net compared against state-of-the-art methods. The results consistently demonstrate the superior performance of our method in room layout estimation. Notably, the model has a novel implementation that enables it to generalize efficiently to cuboid and non-cuboid layouts, solidifying its position as a noteworthy advancement in the field of panorama image understanding. Furthermore, the paper emphasizes the compatibility of the TPS transformation with panoramic images, emphasizing an essential component of the model's effectiveness.%
The model's effectiveness is further supported by its high 3DIoU scores: 85.49 on the PanoContext dataset, 86.16 on Stanford-2D3D, 81.76 on Matterport3DLayout, and 91.98 on ZInD. These results confirm the robustness of the proposed method in addressing the challenging task of panoramic room layout estimation. In conclusion, PanoTPS-Net offers not only a powerful solution for layout prediction but also valuable insights into the compatibility of spatial transformations with panoramic image geometry.

Although the proposed model achieves superior performance in the general room layout estimation task, it has a single limitation. In the non-cuboid room scenarios, there are some cases where one corner exists in front of another corner so that one of them is occluded. In such cases, the proposed method may struggle to differentiate between the two corners, potentially considering them as a single corner.

In our future work, we plan to address the previous issue by developing a more robust model capable of handling occlusion. This may be achieved by designing a two-stage model. In the first stage, a classification network is trained to predict the number of corners in the layout. Based on the classification result, the model should choose one of the predefined reference maps that reflects the number of corners. In the second stage, the model proposed in this paper can be used, with the distinction that the reference map is dynamic and selected based on the first-stage prediction. We believe that this approach may provide a solution for the occlusion problem and provide better room layout estimation results.

It is worth noting that the proposed method, based on the TPS transformer, can be considered as a general framework for various tasks in computer vision. The TPS transformer can be easily utilized in tasks such as facial expression translation, human pose estimation, image registration, and others, following a similar approach as outlined in this paper. This involves using a feature extractor network to extract features and employing the extracted features to predict the transformation parameters of the TPS based on a grid-based differentiable transformation.

\section*{Acknowledgments}
This research was partly funded by the Natural Sciences and Engineering Research Council of Canada (NSERC) and the TMU FOS Postdoctoral Fellowship.

\vfill


\begin{thebibliography}{1}
\bibliographystyle{IEEEtran}
\bibitem{b0}Berenguel-Baeta, B., Bermudez-Cameo, J., \& Guerrero, J. J. (2022). Atlanta scaled layouts from non-central panoramas. Pattern Recognition, 129, 108740.
\bibitem{b1}Jaderberg, M., Simonyan, K., Zisserman, A. \& Others Spatial transformer networks. {\em In Advances of Neural Information Processing Systems (NIPs)}. \textbf{28} (2015).
\bibitem{a5}Hedau, V., Hoiem, D. \& Forsyth, D. Recovering the spatial layout of cluttered rooms. {\em IEEE/CVF International Conference on Computer Vision (ICCV)}. pp. 1849-1856 (2009).
\bibitem{b3}Zhang, Y., Song, S., Tan, P. \& Xiao, J. Panocontext: A whole-room 3d context model for panoramic scene understanding. {\em European Conference on Computer Vision (ECCV)}. pp. 668-686 (2014).
\bibitem{a4} Xu, J., Stenger, B., Kerola, T. \& Tung, T. Pano2cad: Room layout from a single panorama image. {\em Winter Conference on Applications of Computer Vision (WACV)}. pp. 354-362 (2017).
\bibitem{a8}Zou, C., Colburn, A., Shan, Q. \& Hoiem, D. Layoutnet: Reconstructing the 3d room layout from a single rgb image. {\em IEEE/CVF Conference on Computer Vision and Pattern Recognition (CVPR)}. pp. 2051-2059 (2018).
\bibitem{a10}Fernandez-Labrador, C., Facil, J. M., Perez-Yus, A., Demonceaux, C., Civera, J., \& Guerrero, J. J. (2020). Corners for layout: End-to-end layout recovery from 360 images.{\em IEEE Robotics and Automation Letters}, 5(2), 1255-1262.
\bibitem{a3}Sun, C., Hsiao, C., Sun, M. \& Chen, H. Horizonnet: Learning room layout with 1d representation and pano stretch data augmentation. {\em IEEE/CVF Conference on Computer Vision and Pattern Recognition (CVPR)}. pp. 1047-1056 (2019).
\bibitem{a7}Yang, S., Wang, F., Peng, C., Wonka, P., Sun, M. \& Chu, H. Dula-net: A dual-projection network for estimating room layouts from a single rgb panorama. {\em IEEE/CVF Conference on Computer Vision and Pattern Recognition (CVPR)}. pp. 3363-3372 (2019).
\bibitem{a1}Wang, F., Yeh, Y., Sun, M., Chiu, W. \& Tsai, Y. Led2-net: Monocular 360deg layout estimation via differentiable depth rendering. {\em IEEE/CVF Conference on Computer Vision and Pattern Recognition (CVPR)}. pp. 12956-12965 (2021).
\bibitem{a6}Xu, J., Zheng, J., Xu, Y., Tang, R. \& Gao, S. Layout-guided novel view synthesis from a single indoor panorama. {\em IEEE/CVF Conference on Computer Vision and Pattern Recognition (CVPR)}. pp. 16438-16447 (2021).
\bibitem{a8-1}Zou, C., Su, J., Peng, C., Colburn, A., Shan, Q., Wonka, P., Chu, H. \& Hoiem, D. Manhattan Room Layout Reconstruction from a Single 360 Image: A Comparative Study of State-of-the-Art Methods. {\em International Journal Of Computer Vision}. \textbf{129} pp. 1410-1431 (2021).
\bibitem{a8-3} Cruz, S., Hutchcroft, W., Li, Y., Khosravan, N., Boyadzhiev, I., \& Kang, S. B. (2021). Zillow indoor dataset: Annotated floor plans with 360deg panoramas and 3d room layouts. In Proceedings of the IEEE/CVF conference on computer vision and pattern recognition (pp. 2133-2143).
\bibitem{a2}Jiang, Z., Xiang, Z., Xu, J. \& Zhao, M. Lgt-net: Indoor panoramic room layout estimation with geometry-aware transformer network. {\em IEEE/CVF Conference on Computer Vision and Pattern Recognition (CVPR)}. pp. 1654-1663 (2022).
\bibitem{mlc1}Solarte, B., Wu, C. H., Liu, Y. C., Tsai, Y. H., \& Sun, M. (2022). {\em 360-mlc: Multi-view layout consistency for self-training and hyper-parameter tuning}. Advances in Neural Information Processing Systems, 35, 6133-6146.
\bibitem{b4}Zhao, Y., Wen, C., Xue, Z. \& Gao, Y. 3D Room Layout Estimation from a Cubemap of Panorama Image via Deep Manhattan Hough Transform. {\em European Conference on Computer Vision (ECCV)}. pp. 637-654 (2022).
\bibitem{a0}Shen, Z., Zheng, Z., Lin, C., Nie, L., Liao, K. \& Zhao, Y. Disentangling Orthogonal Planes for Indoor Panoramic Room Layout Estimation with Cross-Scale Distortion Awareness. {\em IEEE/CVF Conference on Computer Vision and Pattern Recognition (CVPR)}. pp. 17337-17345 (2023).
\bibitem{b6} Pu, G., Zhao, Y., \& Lian, Z. (2024, December). Pano2Room: Novel View Synthesis from a Single Indoor Panorama. In SIGGRAPH Asia 2024. Conference Papers (pp. 1-11).
\bibitem{b78} Li, K., Zhang, T., Zhong, C., Zhang, Z. \& Wang, G. Robust 3D point clouds classification based on declarative defenders. {\em Neural Computing and Applications}, 37(3), pp.1209-1221 (2025).
\bibitem{a8-2}Chang, A., Dai, A., Funkhouser, T., Halber, M., Niebner, M., Savva, M., Song, S., Zeng, A. \& Zhang, Y. Matterport3D: Learning from RGB-D Data in Indoor Environments. {\em 2017 International Conference On 3D Vision (3DV)}. pp. 667-676 (2017).
\bibitem{a11}Chollet, F. Xception: Deep Learning With Depthwise Separable Convolutions. {\em IEEE/CVF Conference on Computer Vision and Pattern Recognition (CVPR)}. (2017).
\bibitem{a11-1}Ibrahem, H., Salem, A. \& Kang, H. Real-Time Weakly Supervised Object Detection Using Center-of-Features Localization. {\em IEEE Access}. \textbf{9} pp. 38742-38756 (2021). 
\bibitem{a12}Xiao, J., Ehinger, K., Hays, J., Torralba, A. \& Oliva, A. Sun database: Exploring a large collection of scene categories. {\em International Journal Of Computer Vision}. {119} pp. 3-22 (2016).
\bibitem{a13}Pintore, G., Agus, M. \& Gobbetti, E. AtlantaNet: inferring the 3D indoor layout from a single 360 image beyond the Manhattan world assumption. {\em European Conference on Computer Vision (ECCV)}. pp. 432-448 (2020).
\bibitem{a15}He, K., Zhang, X., Ren, S. \& Sun, J. Deep Residual Learning for Image Recognition. {\em IEEE/CVF Conference on Computer Vision and Pattern Recognition (CVPR)}. pp. 770-778 (2016).
\bibitem{a16}He, K., Zhang, X., Ren, S. \& Sun, J. Identity Mappings in Deep Residual Networks. {\em European Conference on Computer Vision (ECCV)}. pp. 630-645 (2016)
\bibitem{a17}Szegedy, C., Vanhoucke, V., Ioffe, S., Shlens, J. \& Wojna, Z. Rethinking the Inception Architecture for Computer Vision. {\em IEEE/CVF Conference on Computer Vision and Pattern Recognition (CVPR)}. pp. 2818-2826 (2016).
\bibitem{a18}Liu, Z., Mao, H., Wu, C., Feichtenhofer, C., Darrell, T. \& Xie, S. A ConvNet for the 2020s. {\em IEEE/CVF Conference on Computer Vision and Pattern Recognition (CVPR)}. pp. 11966-11976 (2022).
\bibitem{a19}Huang, G., Liu, Z., Van Der Maaten, L. \& Weinberger, K. Densely Connected Convolutional Networks. {\em IEEE/CVF Conference on Computer Vision and Pattern Recognition (CVPR)}. pp. 2261-2269 (2017).
\bibitem{a19-1}Sun, C., Sun, M. \& Chen, H. HoHoNet: 360 Indoor Holistic Understanding with Latent Horizontal Features. {\em IEEE/CVF Conference on Computer Vision and Pattern Recognition (CVPR)}. pp. 2573-2582 (2021).
\bibitem{a19-2}Wang, H., Hutchcroft, W., Li, Y., Wan, Z., Boyadzhiev, I., Tian, Y. \& Kang, S. PSMNet: Position-aware Stereo Merging Network for Room Layout Estimation. {\em IEEE/CVF Conference on Computer Vision and Pattern Recognition (CVPR)}. pp. 8606-8615 (2022).
\bibitem{a20}Howard, A., Sandler, M., Chen, B., Wang, W., Chen, L., Tan, M., Chu, G., Vasudevan, V., Zhu, Y., Pang, R., Adam, H. \& Le, Q. Searching for MobileNetV3. {\em IEEE/CVF International Conference on Computer Vision (ICCV)}. pp. 1314-1324 (2019).
\bibitem{a21}Tan, M. \& Le, Q. EfficientNet: Rethinking Model Scaling for Convolutional Neural Networks. {\em International Conference on Machine Learning (ICML)}. pp. 6105-6114 (2019).

\bibitem{a27}Howard, A., Sandler, M., Chen, B., Wang, W., Chen, L., Tan, M., Chu, G., Vasudevan, V., Zhu, Y., Pang, R., Adam, H. \& Le, Q. Searching for MobileNetV3. {\em IEEE/CVF International Conference on Computer Vision (ICCV)}. pp. 1314-1324 (2019).
\bibitem{a29}Kolesnikov, A., Dosovitskiy, A., Weissenborn, D., Heigold, G., Uszkoreit, J., Beyer, L., Minderer, M., Dehghani, M., Houlsby, N., Gelly, S., Unterthiner, T. \& Zhai, X. An Image is Worth 16x16 Words: Transformers for Image Recognition at Scale. {\em International Conference on Learning Representation (ICLR)}, (2021).
\bibitem{a30} Tabata, M., Kurata, K., \& Tamamatsu, J. Shape-Net: Room Layout Estimation from Panoramic Images Robust to Occlusion using Knowledge Distillation with 3D Shapes as Additional Inputs.In IEEE/CVF Conference on Computer Vision Pattern Recognit. Workshops (CVPRW). PP 3552--3561. 2023.

\end{thebibliography}
\end{document}